# The Automatic Training of Rule Bases that Use Numerical Uncertainty Representations *

Richard A. Caruana

Philips Laboratories
North American Philips Corporation
345 Scarborough Road
Briarcliff Manor, NY 10510
914-945-6450

## ABSTRACT

The use of numerical uncertainty representations allows better modeling of some aspects of human evidential reasoning. It also makes knowledge acquisition and system development, test, and modification more difficult.

We propose that where possible, the assignment and/or refinement of rule weights should be performed automatically. We present one approach to performing this training - numerical optimization - and report on the results of some preliminary tests in training rule bases. We also show that truth maintenance can be used to make the training more efficient and ask some epistemological questions raised by training rule weights.

## 1.0 THE NEED FOR TRAINING

As knowledge-based systems attempt to incorporate more of the evidential reasoning capabilities of human experts the adoption of numerical representations for uncertainty and imprecision has become more common. While the use of numerical representations does appear to allow better modeling of some aspects of human evidential reasoning, it also makes knowledge acquisition and system development, test, and modification more difficult.

Experts have difficulty translating their expertise into numerical terms. Almost universally they feel uncomfortable assigning and interpreting numerical weights. Ad hoc uncertainty representations make it impossible to objectively determine what weights should be given to even well understood aspects of the problem. Probability-based representations require experts to specify probabilities that they usually do not know. Moreover, failure of the assumptions required by probabilistic formalisms (e.g. independence) can make the acquired weights invalid in the context of the whole system despite their possible validity in isolation.

Most knowledge engineers admit to the necessity of modifying acquired rule weights until adequate system performance is obtained. Manual tuning is both time consuming and inexact. It is often based on inadequate tests and a relatively subjective "feel" of how the system is performing, and local improvements obtained by tuning one capability of the system are sometimes detrimental to other system capabilities.

The automatic tuning of numerical weights in AI systems is not new [1-9]. Samuel [1,2] employed automatic tuning of coefficients in polynomial evaluation func-tions to effect learning in his checkers programs. In this paper we describe the application of numerical optimization to training rule bases that use numerical uncertainty representations. First, we present the method and discuss its special requirements. Then we introduce the use of truth maintenance to improve the method's computational efficiency. After this we present the results of some preliminary experiments performed on a small classification system. Finally, we discuss some of the limitations of this approach and raise several epistemological questions.

## 2.0 TRAINING AS OPTIMIZATION

Rule bases that incorporate numerical rule weights can be trained by treating the rule weights as parameters defining a many-dimensional space in which the minimum of an error metric (corresponding to maximum performance) for the system is to be found. Search for the minimum in the metric space can be performed by any of the many techniques for numerical optimization [10]. Training performed in this manner requires the availability of a large, representative training set, an objective metric of system performance, and a preexisting rule base, possibly with rule weights already assigned.

Figure 1 shows a basic training algorithm using steepest descent minimization. The algorithm searches the rule weight space in an attempt to find the rule weight assignments that minimize the value of the error metric, a continuous measure of the system's *total* performance over the *entire* training set.

Of course many different minimization techniques could be employed to perform training. Tradeoffs must be made between the number of times the expert system must be evaluated, the speed of convergence, and the sensitivity to local minima [8,10,11] when selecting which optimization method to use. We present an unmodified version of the steepest descent algorithm (one of the simpler and more well known minimization techniques) to avoid detracting from the central issues of this paper.

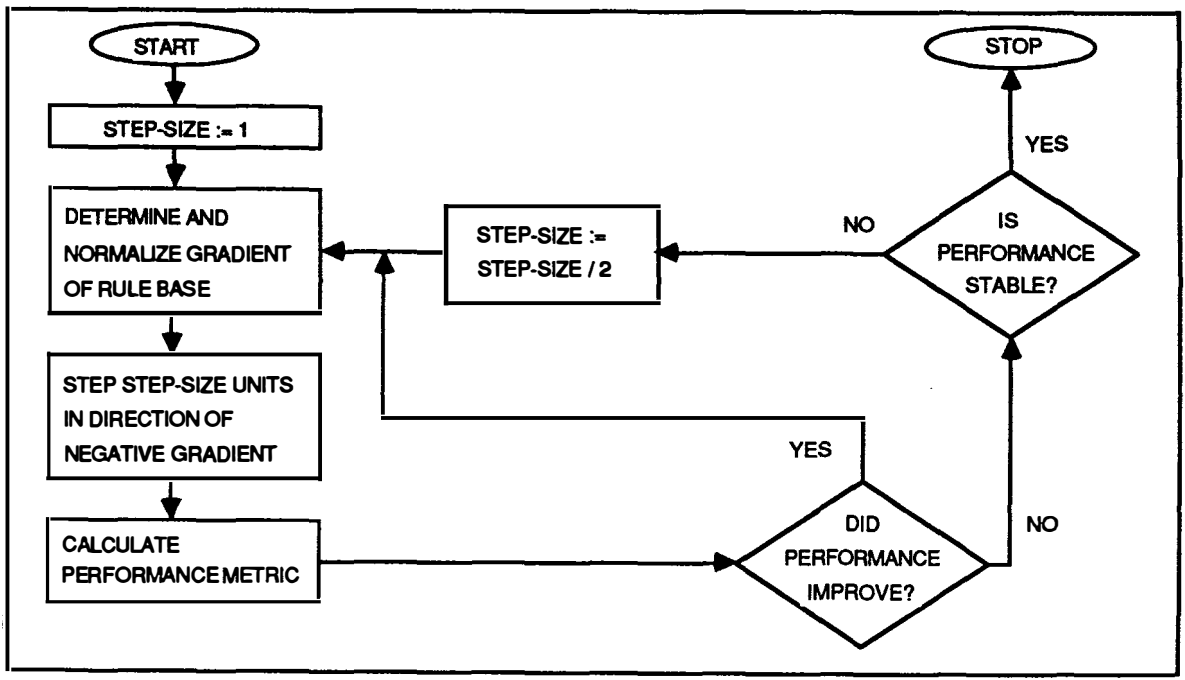

Figure 1. Automatic Training Using Steepest Descent Minimization

## 3.0 IMPROVING EFFICIENCY WITH TRUTH MAINTENANCE

Most minimization algorithms require repeated execution of the rule base during the search process. This is very costly and places upper bounds on the size of the rule base and training data set that can be used. For example, the steepest descent algorithm requires repeated determination of the gradient of the rule-based system. This gradient computation is prohibitive, potentially requiring one complete execution of the rule base for each object in the training set for each rule weight per gradient determination. Thus

$$N = G * O * R \qquad [1]$$

where

$G$ is the number of gradients computed
$O$ is the number of training objects
and $R$ is the number of rules being trained is the total number of evaluations of the rule base required to find the minimum if the search requires $G$ iterations. If the number of rules fired during each execution of the rule base is proportional to the number of rules in the rule base then the total complexity of the minimization is of order $(R^2)$.

Truth maintenance can be used to reduce the cost of the gradient determination. If truth maintenance is employed so that a change in a single rule weight causes only that rule and other rules that depend upon its consequent to refire, then the computational cost of the gradient calculation can be reduced significantly.

The efficiency increase gained by the use of truth maintenance depends upon the structure of the rule base. If the inferencing is very shallow then the resulting complexity is nearly of order $(R)$. Rule bases with inference trees resembling balanced binary trees yield a complexity of order $(R \log R)$. In either case, the efficiency increase resulting from the application of truth maintenance is critical, and without it even very small systems cannot be trained in an acceptable amount of time on the serial hardware currently available.

Truth maintenance is an effective approach to increasing computational efficiency whenever the optimization procedure performs mainly incremental modification to the assigned rule weights. Examples of such procedures are the gradient methods [10] and most simple hill-climbers. Truth maintenance also can be employed to increase the efficiency of many other optimization algorithms that are not explicitly incremental (e.g., simulated annealing [11]) by the use of incremental state change rules.

## 4.0 THE PERFORMANCE METRIC

Many of the more efficient minimization algorithms need a continuous performance metric. This prevents the use of simple metrics that take into account only the number of correct or incorrect conclusions derived by the system. Instead, metrics that determine the degree of rightness or wrongness of the conclusions are required. Fortunately, the conclusions in systems using numerical uncertainty representations typically have continuous confidences associated with them, thereby simplifying the generation of continuous performance metrics. An example of a simple, continuous metric for classification systems is presented in the *Experimental Results* section below.

In many applications the performance metric will include significant domain expertise. For example, in a medical diagnosis expert system it is important not only to determine the most likely cause of the symptoms, but also to asses the risks associated with failure to treat illnesses that are less likely but possibly more dangerous than the most likely diagnosis. Implicit cost functions such as this exist in many domains. Effective training requires that these cost functions be made explicit in the performance metric.

## 5.0 RULE WEIGHT CONSTRAINTS

Many numerical uncertainty representations will require that explicit constraints be placed on the allowable rule weights to prevent optimization from exploring meaningless assignments. Also, it may be desirable for the system builder to be able to explicitly restrict the rule weights that are explored for some of the rules.

Common definitional constraints are ones that restrict the rule weights to the allowable interval, for example [-1, +1] for certainty-factors. Also, for those representations using interval formalisms, constraints can be used to require the lower bound of each rule's confidence interval to be less than or equal to the rule's upper bound.

Additional non-definitional constraints may also be desirable. An expert might wish to restrict the range of a particular rule's weight(s) to some subinterval of the allowable range. For example, restricting rules so that they only provide positive or negative support to their consequent would be common.

Constraints can be represented as penalty functions that make constraint violations appear unattractive or as constraints that are enforced after each iteration has been performed. Care must be exercised in selecting which approach to use and how it is implemented. For example, poorly designed penalty functions can reduce search efficiency and cause unwanted deformations in the metric space and after-iteration enforcement can create "traps" at constraint boundaries. We have used a combination of after-iteration enforcement for theoretic constraints and penalty functions for non-theoretic constraints in our experiments.

## 6.0 EXPERIMENTAL RESULTS

Promising results have been obtained in training experiments with small rule-based systems. For example, a small certainty-factor classification system that was automatically trained outperformed manually generated ones. Moreover, the resulting confidences properly reflected the relationships contained in the training data. Training required about two hours on a Symbolics 3670 using truth maintenance in a hybrid architecture built in LISP and the knowledge engineering tool ART. (A significant speed improvement at the loss of some generality and modifiability could be obtained if the entire system were coded in LISP.) Attempts to train the same system without using truth maintenance were impractical, requiring several hours per iteration and quickly depleting the available virtual memory.

The classification system contained about 50 rules of the form

$$FEATURE_i(x) \rightarrow^{cf} CLASS(x, CLASS_j)$$

where the degree to which $FEATURE_i$ matched each object in the training data was also represented as a certainty-factor. Rules representing each combination of a feature and a class were included. The system explicitly computes the confidence that each object is in each class. The final classification for each object was defined to be that classification for which the system yielded the highest confidence for that object. (Note that this is *not* an ideal application of certainty factors.) In order to make things interesting some of the features selected were irrelevant to the classification problem and the training data represented a very diverse collection of objects from the classes. See [12] for additional detail concerning the rule base and training data.

The performance metric for this system was a simple one appropriate for most certainty factor classification systems where the cost of each misclassification is the same. This metric is:

$$Metric = \sum_i \sum_{j \neq tc(i)} (2 + (CF_{ij} - CF_{i\,tc(i)}))^2 \quad [2]$$

where

    $i$ ranges over all training objects
    $j$ ranges over all classifications
    $CF_{ij}$ is the confidence object $i$ is in class $j$
   and $tc(i)$ is a function that returns the true classification of object $i$

This metric takes into account both the correctness and sharpness of the classification of the system and assumes that the cost of all misclassifications is the same. As described above, more complex metrics

201

would be needed for many domains.

Three different training tests were performed. In the first, all rules were assigned initial confidences of zero. In the second, rules were assigned initial confidences by an expert but the expert was prevented from performing iterative refinement. In the third, rules were assigned initial confidences by the expert and the expert was allowed to perform iterative refinement. Optimisation was performed starting from each set of assigned rule weights.

All of the trained expert systems (including the one initialised with zero weights) outperformed the best manually generated system in both classification accuracy and in the sharpness of the classifications. In fact, all of the trained systems exhibited performance that the expert had considered unattainable given the complexity of the training data and the simplicity of the rule base. Although all three of the automatically trained systems outperformed the best manually generated system they did not yield identical performance or result in identical rule weights. This is due to the minimisation routine getting stuck in local minima. The tuned system that performed the best was the one that was started from the best human estimates of the rule weights.

## 7.0 DISCUSSION

The sensitivity of rule bases to the modification of rule weights has been discussed briefly by Buchanan and Shortliffe [13]. Their conclusion, based upon tests with the MYCIN system, was that the performance of the system was relatively insensitive to the precision of the weights. Their conclusion, however, is probably not broadly applicable because of characteristics particular to their domain. Specifically, the MYCIN system derives the set of most likely hypotheses rather than having to select the single best one; the final recommended treatment is relatively insensitive to variations in the confidences associated with each hypothesis; and MYCIN's inferencing is relatively flat. Moreover, the data they present, contrary to their interpretation, does seem to indicate a significant sensitivity to the precision of the confidences. Given that the MYCIN experiment altered only the precision of the rule weights and did not change the relative ranking of rules, we believe that the results they report do not conflict with those we present here.

Training need not be performed on all of the rule weights in the system. It can be performed on only a specific subset of the rule base at any one time. This could be implemented by utilising constraints, but this would be inefficient. The approach we have used is to utilise a list of rules that are to be optimised. Rules not contained in this list can fire but are not optimised. This allows system modules to be trained independently provided that a suitable metric of that module's performance is available. We have not yet attempted to train separate system modules and therefore do not know if this represents a viable approach to handling scale-up problems.

The success of training via optimisation depends critically upon the quality of the performance metric. In some domains the performance metric could easily be as complex as the expert system and incorporate significant domain expertise. Although this may be a costly addition to the system development process, we believe objective performance criteria are critical to the success of any AI development effort. Given that final performance is one of the primary ways the end user will judge the system, this emphasis on performance can be advantageous.

This approach requires the availability of a large, representative training set. A large training set may not be available in some domains and may not be easy to generate. In these domains traditional knowledge engineering methods are likely to prevail, though the need to thoroughly test the resulting systems in the absence of a large body of test data still poses a significant problem. As with any training algorithm there is the possibility of overtraining. Reduced performance on data not used for training may indicate that over-training has occurred. More work needs to be done to better define the characteristics of a good training set and to determine how to detect and possibly minimize the effects of overtraining.

Local minima may present problems when training. Their effects on the final performance of the trained system can be reduced by giving the system a reasonable initial parameterization and by starting from several different initial parameterizations. The avoidance of local minima is an active area of research in optimization. Most advances in numerical optimization would be applicable to the training procedure we describe. Optimization by simulated annealing [7] looks particularly attractive because of its relative immunity to local minima but we have not yet tested it.

## 8.0 EPISTEMOLOGICAL ISSUES

There are epistemological questions raised by the automatic training of rule weights. The weights assigned by an expert may not be optimal, but at least they reflect an expert's understanding of the domain. Training implies that not all of the expertise in the final system is derived from the expert.

In a trained system only the symbolic part of the rule and not the rule's actual power is derived from the expert. For example, it is possible that training would reduce the weight of some rule to zero, effectively yielding a rule base equivalent to one where that rule has been deleted. Thus training has the power to reject expertise added by the expert. Of course, rule weights acquired from an expert should not be preferred over automatically acquired weights that yield significantly better performance, but care must be taken to insure that training does not actually reduce performance in cases considered by the expert but not thoroughly represented by the training data or improperly reflected in the performance metric. Training, therefore, has implications for the ultimate accountability and trust in the final system.

Another question is what interpretation is to be given to trained rule weights. If the formalism is ad hoc, then no interpretation other than relative strength need be given. If, however, the formalism has a probabilistic interpretation, then one would expect the assigned weights to properly reflect the probabilities represented in the training data. In fact, here one might dispense with optimization altogether and instead directly compute the weights from the training data. But failure of independence assumptions or an inaccurate or incomplete rule base could cause the computed rule weights to differ significantly from the optimally performing ones. Thus optimal rule weights might differ significantly from what is expected. Nonetheless, if the system is to provide meaningful explanations or is expected to represent a useful encapsulation of human expertise then it would be desirable to have training that yields rule weights "useful" in contexts beyond their performance within the trained system.

## 9.0 SUMMARY

An approach to training rule-based systems that incorporate numerical uncertainties via numerical optimization has been introduced. The use of truth maintenance to increase the efficiency of this approach has been presented. The computational cost of the training process is high. Nevertheless, compared with the time that it takes experts and knowledge engineers to perform similar and possibly less effective training, this is probably a time-efficient approach to system development in some domains.

Tests performed on a simple certainty-factor classification system indicate that the technique may be a viable supplement to knowledge acquisition and maintenance techniques. Further testing remains to be done to determine if the technique will scale-up.

Epistemological issues concerning accountability and the interpretation of automatically acquired rule weights were also raised.